\pdfoutput=1

\documentclass[11pt]{article}

\usepackage[]{acl}

\usepackage{times}
\usepackage{latexsym}

\usepackage[T1]{fontenc}

\usepackage[utf8]{inputenc}

\usepackage{microtype}
\usepackage{enumitem}
\usepackage{amssymb}
\usepackage{mathrsfs}
\usepackage{amsmath}
\usepackage{hhline}
\usepackage{amsfonts}
\usepackage{multirow}
\usepackage{array}
\usepackage{booktabs}
\usepackage{colortbl}
\usepackage{color}
\usepackage{subcaption}
\usepackage{threeparttable}
\usepackage{float}
\usepackage{caption}
\usepackage{graphicx}
\usepackage{arydshln}
\usepackage{makecell}
\usepackage{inconsolata}

\definecolor{mygray}{gray}{0.9}
\definecolor{mypink}{rgb}{0.99,0.91,0.95}
\definecolor{mycyan}{cmyk}{0.3,0,0,0}
\title{Incorporating Distributions of Discourse Structure for Long Document Abstractive Summarization}

\author{Dongqi Liu \quad Yifan Wang \quad Vera Demberg \\
        Department of Computer Science\\
        Department of Language Science and Technology\\
        Saarland Informatics Campus, Saarland University, Germany\\
        \texttt{\{dongqi,yifwang,vera\}@lst.uni-saarland.de}}

\begin{document}
\maketitle
\begin{abstract}
For text summarization, the role of discourse structure is pivotal in discerning the core content of a text. Regrettably, prior studies on incorporating Rhetorical Structure Theory (RST) into transformer-based summarization models only consider the nuclearity annotation, thereby overlooking the variety of discourse relation types. This paper introduces the `RSTformer', a novel summarization model that comprehensively incorporates both the types and uncertainty of rhetorical relations. Our RST-attention mechanism, rooted in document-level rhetorical structure, is an extension of the recently devised Longformer framework. Through rigorous evaluation, the model proposed herein exhibits significant superiority over state-of-the-art models, as evidenced by its notable performance on several automatic metrics and human evaluation.\footnote{The project information can be accessed by visiting: \href{https://dongqi.me/projects/RSTformer}{https://dongqi.me/projects/RSTformer}.}
\end{abstract}

\section{Introduction}
For writing a good summary of a long document, it is of paramount importance to discern the salient information within the text and to comprehend the intricate interconnections among its various components. Contemporary leading-edge systems for abstractive (long) text summarization employ Transformer \cite{NIPS2017_3f5ee243} encoder-decoder architecture \citep{NEURIPS2020_c8512d14, guo-etal-2022-longt5}. These sequence-to-sequence (seq2seq) models first transform the source document into a high-dimensional content representation and then decode the predicted summary conditioned on the representation \citep{belinkov2018synthetic, xu-durrett-2019-neural, cao-wang-2022-hibrids, balachandran-etal-2021-structsum}. It has been demonstrated in the past that such an architecture does a poor job of digging high-level discourse structure during the encoding phase \citep{lin-etal-2019-open, zhang2020sg, koto-etal-2021-discourse, de2023evaluation}. However, discourse structure is very important for deciding what to include vs.~not to include in the summary \citep{marcu1997discourse, marcu1999discourse, marcu1998improving, zhong2020discourse}. Given that previous work has indicated that the performance of neural language models can be enhanced through the incorporation of latent structure information \citep{ettinger2020bert, miaschi-etal-2020-linguistic, qian-etal-2021-structural, pu-simaan-2022-passing}, we will here explore the integration of discourse relation structure into the Longformer model \citep{beltagy2020longformer}; this architecture has been shown to be particularly suitable for encoding long input texts.

\begin{figure}[t]
\centering
\includegraphics[width=0.3\textwidth]{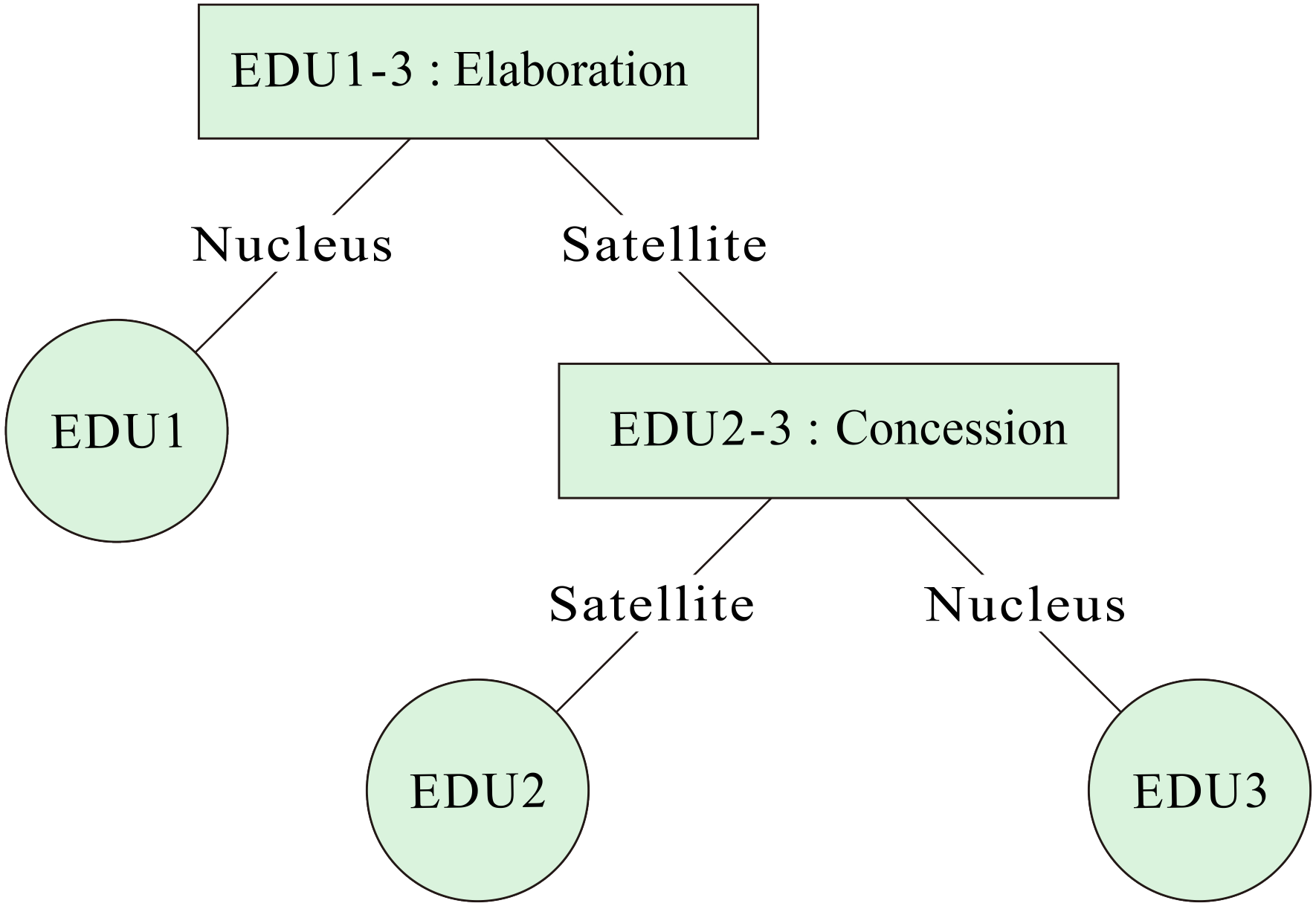}
\captionsetup{belowskip=-0.15cm}
\caption{An example of RST tree: [\textit{Rhetorical structure theory (RST) is a theory of text organization.}]$^{\mathrm{EDU1}}$ [\textit{Although the RST structure is difficult to annotate,}]$^{\mathrm{EDU2}}$ [\textit{there are still many scholars who have studied it.}]$^{\mathrm{EDU3}}$}
\label{fig:rst}
\end{figure}

Rhetorical Structure Theory (RST) serves as a discourse framework designed to articulate the interrelationships among sentences at the document level. This framework distinguishes a plethora of coherence relations delineating the manner in which two text segments are interconnected (e.g., one segment might give a reason for a claim made in another segment, or alternatively, two segments may contrast with one another). Moreover, RST distinguishes between paratactic relations, where both segments carry equivalent discourse importance, and hypotactic relations, which classify the segment of greater centrality to the overarching discourse structure as the `nucleus' and the less central one as the `satellite'. Figure \ref{fig:rst} shows a simple example of an RST tree. In this instance, EDU1 serves as the nucleus of the elaboration relation, whereas the combination of EDUs 2 and 3 constitutes the satellite of said relation. Furthermore, we can see that EDU3 assumes a more central role within the concession relation, hence it is marked as its nucleus, while EDU2 holds less important: if EDU2 was left out, the elaboration relation between EDUs 1 and 3 would still hold, but if EDU3 was removed, an elaboration relation between EDU1 and EDU2 would not hold, and the coherence would be lost. As has been recognized early on \citep{marcu1997discourse, marcu1999discourse}, this discourse information can be effectively used in summarization tasks.  

While there have been some previous attempts at integrating discourse structure into neural text summarization models, as seen in \citet{gabriel-etal-2021-discourse, dong-etal-2021-discourse, xiao-etal-2020-really, xu-etal-2020-discourse, cohan-etal-2018-discourse}, these approaches do not utilize relation labels and solely consider the 1-best RST tree obtained from preprocessing of a discourse parser. We argue that this leads to two significant issues: Firstly, information pertaining to relation type is overlooked, despite its known relevance to the summarization task. Secondly, there may be benefits in considering distributions over coherence relation labels, rather than limiting analysis to the 1-best results \cite{pu-simaan-2022-passing}. One reason is that external discourse parsers are known to perform poorly on out-of-domain data \cite{atwell-etal-2022-change, liu-etal-2021-dmrst, gessler-etal-2021-discodisco, koto-etal-2021-top, liu-etal-2020-multilingual-neural, nguyen-etal-2021-rst}, and may hence propagate errors into the summarization model. There is a subsequent risk that these errors will be incrementally amplified during back-propagation, thus potentially impairing the model's performance. A second reason is that there might inherently be several coherence relations holding at the same time \citep{yung-etal-2022-label}, which might be beneficial to represent through the distributions of the discourse structure. Hence, we posit that the output of the RST parser holds greater significance when it not only provides the model with the n-best results but also conveys the remaining uncertainty associated with them.

In the remainder of the paper, we explore whether incorporating the labeled discourse relation structure with uncertainty, which can be understood as the distributions of discourse structure, into the attention mechanism can effectively augment the performance of neural summarization models. Our main contributions are as follows:

\begin{itemize}[itemsep=1pt,topsep=1pt,parsep=1pt]
    \item We represent a generic approach for infusing labeled discourse relations with uncertainty into the encoder’s self-attention layer of Longformer, wherein the self-attention heads are made to specialize in specific discourse categories. Additionally, our modules are orthogonal to the choice of the underlying encoder-decoder Transformer-based architecture, thereby enabling them to be seamlessly incorporated into other advanced models.
    \item We provide empirical evidence supporting the notion that conveying uncertainty and introducing labeled discourse relations to the Transformer are complementary actions, both significantly contributing to the enhancement of the final performance. Our model also surpasses current state-of-the-art models across multiple evaluation metrics.
    \item Quantitative and qualitative analyses show that our model exceeds the baseline model in both novel word generation and factual consistency checking. Furthermore, our model comes closer to human answers in terms of sentence alignment and overall generation quality.
\end{itemize}

\section{Related Work}

\subsection{Text Summarization with RST}

Rhetorical Structure Theory offers a structured paradigm for describing how various discourse units relate to one another in a text. The RST tree structure, as illustrated in \citet{marcu1997discourse} and \citet{louis-etal-2010-discourse}, can serve as a valuable tool for content selection in the process of summarization.

For instance, \citet{kikuchi-etal-2014-single} characterize the dependencies between sentences by constructing RST trees and pruning the parts that are marked as `satellites' while preserving the important content (`nucleus') of the document as predicted summaries. Although RNN-based models are sometimes argued to be sufficient in implicitly learning discourse and semantic relations, \citet{liu-etal-2019-single}'s work underscores the value of explicitly integrating RST trees into the summarization model, thereby highlighting the significance of discourse relation for the neural summarization network. It is also worth noting that while the attention mechanism can more effectively uncover discourse relations without explicit training, it tends to unearth only superficial discourse structure and is often prone to mistakes \cite{vig-belinkov-2019-analyzing, sachan-etal-2021-syntax, xiao-etal-2021-predicting, huber2022towards, davis-van-schijndel-2020-discourse}. 

Although attention-based models excel in executing downstream tasks such as summarization, the explicit incorporation of discourse relations can yield additional benefits. Work highly related to ours includes the model of \citet{xiao-etal-2020-really}, which improves the performance of an extractive summarization model by transmuting the RST structure into a dependency tree and explicitly integrating it into the computation of the attention mechanism. Follow-up works \citet{xu-etal-2020-discourse} and \citet{dong-etal-2021-discourse} further confirm the influence of RST structure on improving attention mechanism by incorporating discourse structure into a transformer-based model and a graph neural network model for the summarization task, respectively. However, all of these neural strategies apply the one-best structure derived from an external discourse parser.

\subsection{Text Summarization with Longformer}

The Longformer model \cite{beltagy2020longformer}, based on a sparse attention mechanism, is considered to be an effective means for processing long documents. Its essence is to make each token only pay attention to a window of a certain size, so that the time complexity of the model is reduced from a quadratic correlation with the text length to a linear correlation. Longformer-related models have since been employed in several summarization tasks \citep[e.g.,][]{zhang-etal-2022-summn, otmakhova-etal-2022-patient, elaraby-litman-2022-arglegalsumm, xie-etal-2022-gretel, pu-etal-2022-two}.

At the same time, there have also been recent attempts at integrating text structure information with the Longformer model in summarization tasks. \citet{huang-kurohashi-2021-extractive} first employ the Longformer to encode input documents and propose an extractive summarization model based on a heterogeneous graph of discourse and coreference relations. \citet{liu-etal-2021-hetformer} extend the Longformer to model different types of semantic nodes in the original text as heterogeneous graphs and directly learn relations between nodes. Specifically, they treated tokens, entities, and sentences as different types of nodes, and the multiple sparse masks as different types of edges to represent relations (e.g., token-to-token, token-to-sentence). \citet{elaraby-litman-2022-arglegalsumm} improve the performance of the strong baseline Longformer by integrating argument role labeling into the summarization process to capture the argumentative structure of legal documents. \citet{ruan-etal-2022-histruct} and \citet{cao-wang-2022-hibrids} enhance extractive and abstractive summarization tasks, respectively, by introducing the text's hierarchical structure (e.g., section title) into the Longformer model.

\section{Proposed Approach}

In the realm of document discourse parsing, the performance of the RST parser leaves much to be desired \cite{yu-etal-2022-rst, nguyen-etal-2021-rst, liu-etal-2021-dmrst}, with parsing performance deteriorating in conjunction with escalating document complexity. Merely passing the 1-best RST tree risks imparting misleading information to the summarization model.

Inspired by \citet{pu-simaan-2022-passing}, the approach to alleviating the aforementioned problems is that we retain uncertainty inside the parser, which can convey the parser's confidence in each discourse relation. Furthermore, we contend that discourse relation labels (types) can provide more fine-grained labeled probability distributions that can assist attention heads of the Transformer-based model to capture the importance of different discourse units. This in turn would contribute to a more precise estimation of the context vector and can enhance the quality of source document encoding. Discourse parsers tend to be more precise (and have more peaked probability distributions) for local coherence relations, which span only a short amount of text, compared to global relations spanning large portions of a text. This aligns well with the dilated (yet still limited) sliding window attention mechanism of the Longformer \cite{beltagy2020longformer}. We, therefore, integrate the probability distributions over local coherence relations into the attention window $w$ of the Longformer.

\subsection{RST Tensor with Labeled Distributions}
The discourse-driven neural seq2seq summarization task can be modeled as follows:

\begin{equation}
\begin{aligned}
    P(t|s,d) & \approx \prod_{i=1}^{T} P(t_i|t_{<i},\mathrm{encode(s},d)) \\
\end{aligned}
\end{equation}

In the above equation, $s$, $t$, and $d$ denote the source, target sequence, and discourse representation, respectively. $T$ signifies the target sequence length and $\mathrm{encode(\cdot)}$ represents the encoder of the summarization model. Previous research \cite{xu-etal-2020-discourse, cohan-etal-2018-discourse, dong-etal-2021-discourse, li-etal-2020-composing, chen-yang-2021-structure} has confirmed that the probability of generating appropriate summaries by incorporating $d$ into the model's encoder is significantly greater than the probability of generating proper summaries without the incorporation of $d$.

Our main idea is to find a better method to incorporate discourse structure $d$. To inject discourse structure, we first apply a `matrixization' approach to represent the discourse structure and produce a compact tensor representation appropriate for the Longformer model \cite{pu-simaan-2022-passing}.

\begin{figure}[t]
\centering
\includegraphics[width=0.35\textwidth,height=0.55\textwidth]{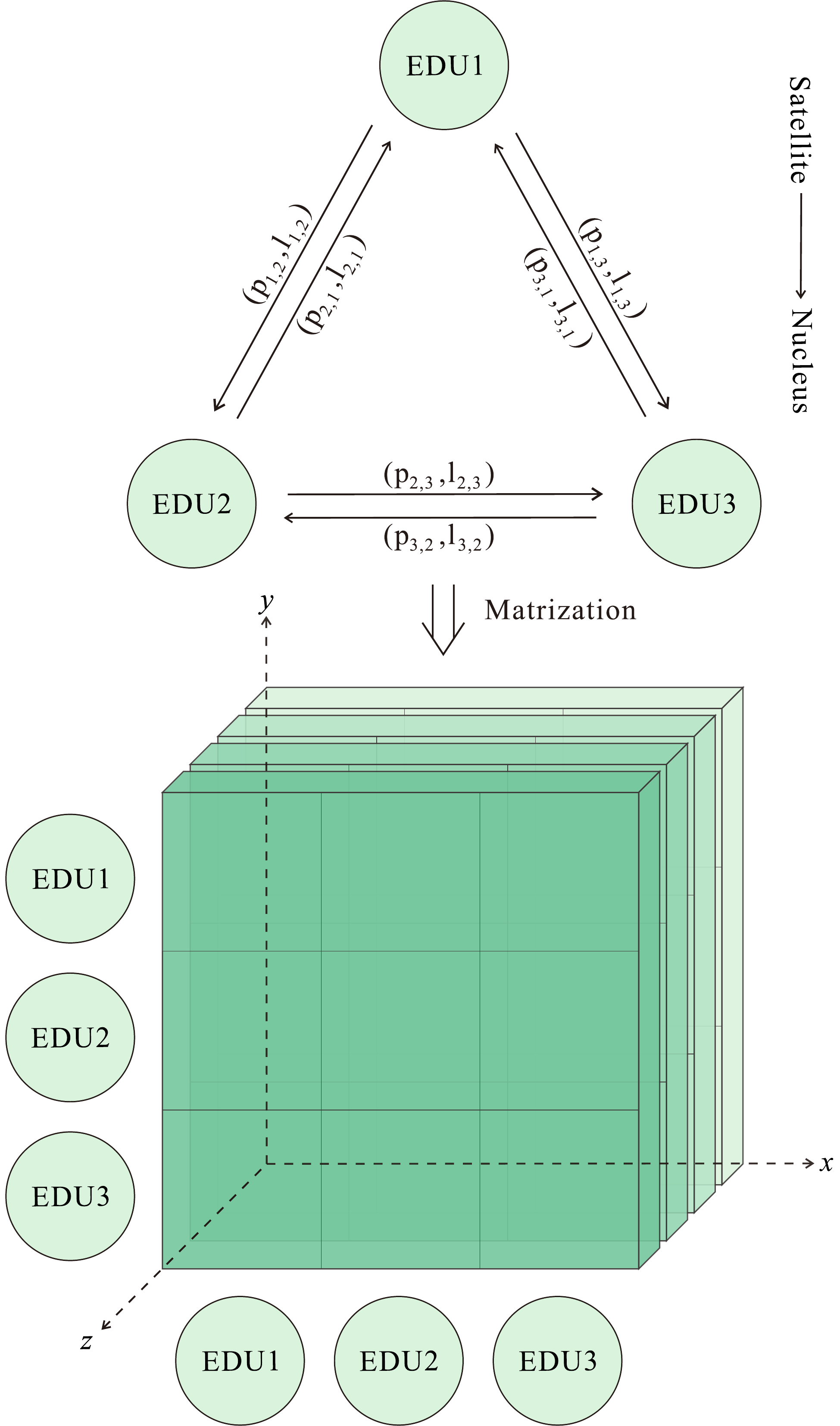}
\caption{Labeled discourse distributions}
\label{fig:labeled discourse distributions}
\end{figure}

Figure \ref{fig:labeled discourse distributions} illustrates by an example how we convert the graph of all potential RST relations (including the n-best RST trees present within the graph) into a three-dimensional labeled discourse distribution (LDD) tensor. The x-axis and y-axis of the tensor represent the elementary discourse units (EDUs) in the source document, while the z-axis represents the type of discourse relation. Each point represents a confidence value $p(edu_i,edu_j,l) \in [0,1] \subseteq \mathbb{R}$ of an elementary discourse unit $edu_i$ connecting to another elementary discourse unit $edu_j$ from source text via the relation $l$. It should be noted that the generation of the LDD tensor should meet the conditions: 1) $p(edu_i,edu_i)=0$, as no unit is dependent on itself; 2) we only extract the relation probability of nucleus units, since nucleus EDUs are more central to the text and should be given more attention. In the example shown in Figure \ref{fig:rst}, we only extract the discourse relation probabilities of EDU1 and EDU3.

\subsection{RST Sparse Attention}

\begin{figure*}[t]
\centering
\includegraphics[width=1\textwidth,height=0.55\textwidth]{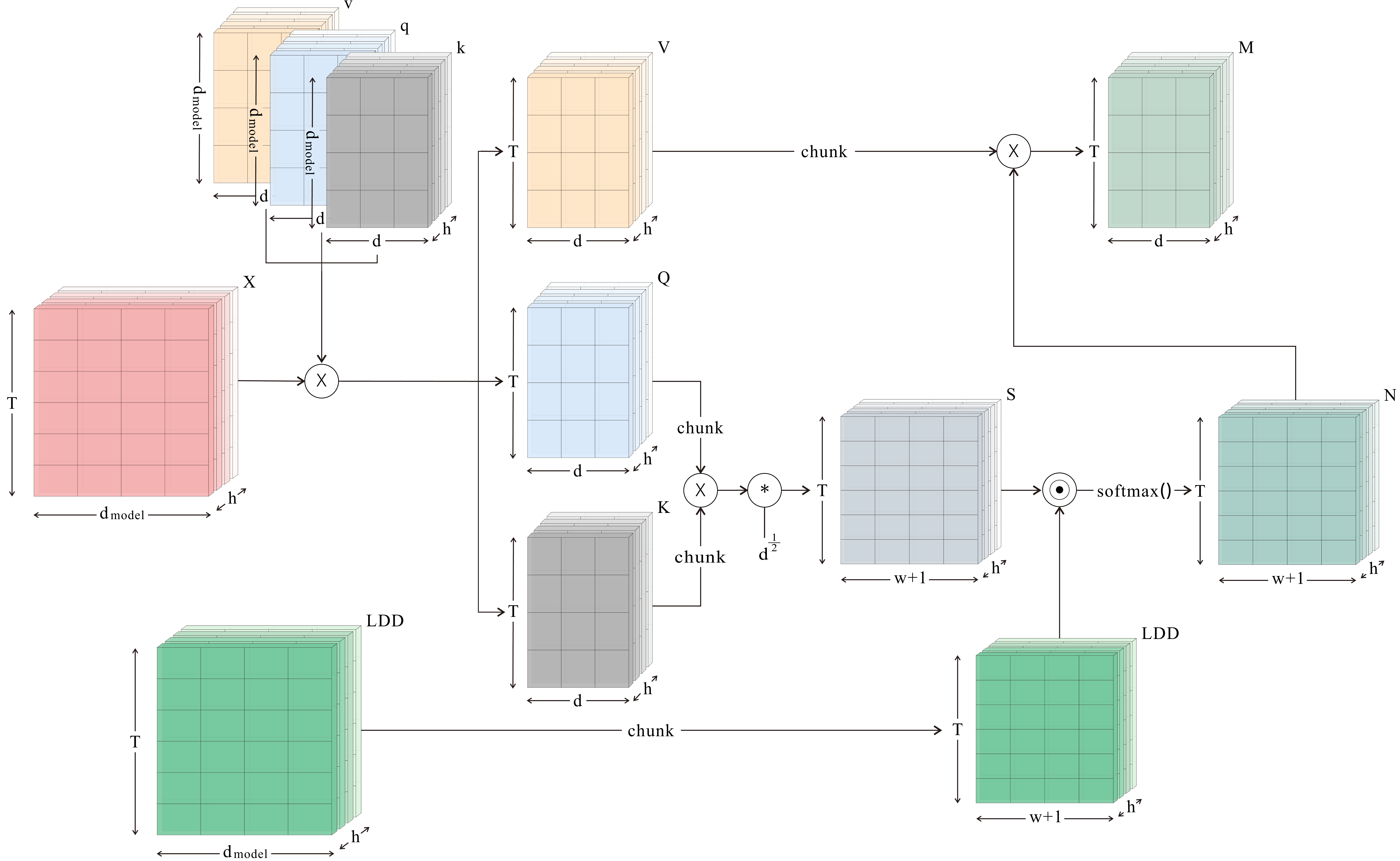}
\caption{Model architecture: we show a schematic diagram of incorporating $\mathrm{LDD}$ tensor into the attention layer of the model. Specifically, $\mathrm{X}$ is text embedding matrix, and $\mathrm{LDD}$ is incorporated with attention matrix $\mathrm{S}$ in the form of element-wise multiplication. In order to ensure the consistency of matrix shape, we also apply an identical \textit{chunk} method as Longformer in $\mathrm{LDD}$.} 
\label{fig:model}
\end{figure*}

We propose a novel Longformer-Encoder-Decoder Summarization model: RSTformer, which incorporates LDD into each layer of the Longformer encoder in a discourse-aware manner. Given that each encoder layer shares an identical configuration, Figure \ref{fig:model} displays one layer architecture of our proposed model.

The standard dilated sliding window attention layer of Longformer employs a multi-head fixed-size window attention mechanism. For a pre-specified window size $w$, each token attends to $\frac{1}{2}w$ tokens on either side. For an input sequence of length $T$, the input of dilated sliding window attention heads in the RSTformer layer comprises the hidden representation tensor $\mathrm{X} \in \mathbb{R}^{T \times d_{model} \times h}$ and labeled discourse distribution tensor $\mathrm{LDD} \in \mathbb{R}^{T \times d_{model} \times h}$, where $d_{model}$ represents the size of the hidden representation and $h$ denotes the number of attention heads. 

As usual in multi-head self-attention, we multiply the text feature representation tensor with $\mathrm{q},\mathrm{k},\mathrm{v} \in \mathbb{R}^{d_{model} \times d \times h}$ to obtain the corresponding $\mathrm{Q} \in \mathbb{R}^{T \times d \times h}$, $\mathrm{K} \in \mathbb{R}^{ T \times d \times h}$, and $\mathrm{V} \in \mathbb{R}^{T \times d \times h}$ matrices, where $d=d_{model}/h$. Subsequently, the attention weight matrix is obtained by:

\begin{equation}
     \mathrm{S} = \frac{\mathrm{Q} \cdot \mathrm{K}^{\top}}{\sqrt{d}}
\end{equation}

Longformer utilizes two sets of projections, $\mathrm{Q_s}$, $\mathrm{K_s}$, $\mathrm{V_s}$ to compute the attention scores of sliding window attention, and $\mathrm{Q_g}$, $\mathrm{K_g}$, $\mathrm{V_g}$ to compute attention scores for global attention. Notably, $\mathrm{Q_g}$, $\mathrm{K_g}$, $\mathrm{V_g}$ are all initialized with values that match $\mathrm{Q_s}$, $\mathrm{K_s}$, $\mathrm{V_s}$ respectively. The dilated sliding window attention operates by calculating a fixed number of the diagonals of $\mathrm{QK^{\top}}$ through sliding chunks query-key multiplication. This process yields a resulting tensor $\mathrm{S} \in \mathbb{R}^{ T \times w+1 \times h}$. Similarly, $\mathrm{LDD}$ and $\mathrm{V}$ adopt the same \textit{chunk} method as employed by Longformer to acquire the sliding window attention matrix.

It should be noted here that we inject the sliding window attention tensor $\mathrm{S}$ obtained from the preceding computation by element-wise multiplication with the LDD tensor:

\begin{equation}
    \mathrm{S} \odot \mathrm{LDD}
\end{equation}

The motivation behind employing element-wise multiplication is to allow the learning parameters of the attention mechanism `dynamically' to optimize the summarization objective but also diverge the least from the parser probabilities in discourse distribution \cite{pu-simaan-2022-passing}. The estimation of attention weights is adjusted to align with the utility of discourse relations for the ultimate summarization task.

Following, the obtained weights are further processed using the softmax function to derive the final tensor representing the discourse-infused distribution:

\begin{equation}
 \mathrm{N} = \mathrm{softmax}(\mathrm{S} \odot \mathrm{LDD})
\end{equation}

It should be emphasized that each attention head is assigned a different discourse matrix $\mathrm{LDD}_l$ for a specific relation $l$, This allocation enables heads to concentrate on and learn different discourse labels \cite{pu-simaan-2022-passing}. In doing so, attention heads can be specialized and acquire a deeper understanding of the impact of discourse labels.\footnote{Appendix \ref{sec:appendixA} details the grouping of discourse relations.}

Finally, the discourse-injected weights $\mathrm{N}$ are multiplied with the value matrix $\mathrm{V}$ to obtain the attention weights $\mathrm{M}$ for this layer and then transfer $\mathrm{M}$ to the next Longformer encoder layer for further computation.

\begin{equation}
    \mathrm{M} = \mathrm{N} \cdot \mathrm{V}
\end{equation}

\section{Experiments and Analysis}

\subsection{Experimental Setup}

\paragraph{Parser} We employ an external RST parser called {\em DMRST}\/ \cite{liu-etal-2021-dmrst, liu-etal-2020-multilingual-neural} to automatically parse the source documents. The probability or uncertainty of discourse relations is extracted from the logits layer of the DMRST\footnote{https://github.com/seq-to-mind/DMRST\_Parser} model. In cases where DMRST fails to parse the source document, we simply skip the LDD generation process and proceed with the normal Longformer procedure.

\paragraph{Datasets} We conduct our experiments on three recent long document summarization datasets: BookSum Chapter \cite{kryscinski-etal-2022-booksum}, eLife \cite{goldsack-etal-2022-making}, and Multi-LexSum \cite{shen2022multi}. We choose these datasets because of their high heterogeneity and we want to investigate whether our approach can maintain adequate generalization performance across different data domains. Table \ref{tab:Datasets discription} shows the statistics of the datasets.

\begin{table*}[ht]
\centering
\scalebox{0.8}{
\tabcolsep=3pt
\begin{tabular}{c c c c c c c c c}
\toprule
Dataset & Training & Validation & Test & Avg. Doc Words & Avg. Summary Words & Coverage & Density & Compression Ratio\\
\midrule
BookSum Chapter & 9600 & 1431 & 1484 & 3834.40 & 363.81 & 0.764 & 1.504 & 15.198 \\
eLife & 4346 & 241 & 241 & 10133.07 & 382.69 & 0.819 & 1.761 & 27.650 \\
Multi-LexSum & 3177	& 454 & 908	& 58210.99 & 547.04 & 0.926 & 3.394 & 95.390\\
\bottomrule
\end{tabular}
}
\caption{Datasets statistics}
\label{tab:Datasets discription}
\end{table*}

\textit{Coverage} refers to the percentage of words in the summary that are from the source document. A higher coverage ratio indicates that a greater proportion of summary words are derived directly from the source text. It is mainly used to measure the degree of derivation of the summary from the text. \textit{Density} is defined as the average length of the extracted segments to which each summary word belongs \cite{segarra-soriano-etal-2022-dacsa}. \textit{Compression ratio} is defined as the ratio between the length of the source document and summary \cite{scialom-etal-2020-mlsum}.

\paragraph{Evaluation Metrics} We evaluate the quality of different summarization systems using Rouge-\{1, 2, L\} score \cite{lin-2004-rouge}, BERTscore \cite{zhang2019bertscore}, Meteor score \cite{banerjee-lavie-2005-meteor}, \{1, 2, 3, 4\}-gram novelty \cite{kryscinski-etal-2018-improving}, SummaC \cite{laban-etal-2022-summac} and sentence alignment \cite{liu-liu-2021-simcls} as criteria for the model’s effectiveness.

In detail, Rouge-\{1,2\} is mainly evaluated based on the co-occurrence of \{1,2\}-gram in summary, while the calculation of Rouge-L uses the longest common subsequence. BERTScore is used to compute the semantic similarity score of candidate sentences to reference sentences through contextual embedding. Meteor is an improvement based on BLEU \cite{papineni-etal-2002-bleu}, which also considers the impact of sentence fluency and synonyms on semantics. \{1, 2, 3, 4\}-gram novelty indicates the capacity of the model to generate new words, rather than merely extracting words from the original text. SummaC detects semantic inconsistency by segmenting documents into sentence units and aggregating scores between sentence pairs.

\paragraph{Training and Inference}

Hyper-parameters for the baseline, proposal models, and ablation models are all kept identical. We adopt the same configuration as Longformer \cite{beltagy2020longformer}: All experiments are optimized using Adam \cite{kingma2014adam} ($\beta_1$ = $0.9$, $\beta_2$ = $0.999$, $\epsilon$ = $10^{-9}$, and weight decay = $0.1$) with Adafactor \cite{shazeer2018adafactor}, the number of warm-up steps is $1500$, and the initial learning rate is set to $3e^{-9}$ with cosine learning rate schedule. We also apply NoisyTune (Noise lambda = $0.2$) \cite{wu-etal-2022-noisytune} for efficient fine-tuning. The size of the local attention window is $w$ = $1024$, and we choose cross-entropy as loss function. 

During the training phase, we save the checkpoint with the highest Rouge-2 F1 score on the validation set as the final model. The experiments are all run for $30$ epochs using a batch size of $1$ with early stopping implemented. In order to prevent over-fitting, we set the dropout rate to $0.1$ in all layers of the model. For model inference, we adopt a beam size of $4$ with a length penalty of $2.0$, and we set the no-repeat n-gram size to $3$. 

\subsection{Results}

The experimental results for each model are presented in Table \ref{tab:Model Performance}. To estimate a lower bound in performance, we simply use the original document as the summary. Further trivial models include the Lead-3 model which simply picks the first three sentences of the document as the summary. Lead-K similarly extracts the first K sentences of the document, until a similar length as the reference summary is reached. Longformer and state-of-the-art (SOTA) models serve as our baseline and comparison models, respectively. The remaining two models are the models we proposed. RSTformer (w/o relations) refers to the model that preserves whether there are relations between EDUs and ignores the type of relations by summing the third dimension of LDD tensors. RSTformer (w relations) is the final model we propose, with the only difference being the inclusion of the impact of RST types.

Both RSTformer versions are found to outperform the baseline model on various measures. The higher scores reflect an improved choice of words (Rouge \& Meteor scores), and also the semantics of the text (BERTscore).\footnote{The version of BERTscore we use comes from the original paper version \cite{zhang2019bertscore} with HuggingFace default API (https://huggingface.co/spaces/evaluate-metric/bertscore).} The proposed model, RSTformer, demonstrates robust generalization capabilities across different datasets, highlighting its promising potential in various summarization domains.

In most of our summarization experiments, we furthermore find that incorporating discourse structure with types provides better experimental results than the discourse distributions without types, even beating the SOTA model on our experimental datasets. This observation suggests that providing more discourse information, especially type distribution probabilities, is a promising approach.

\begin{table*}[ht]
\centering
\scalebox{0.79}{
\tabcolsep=3pt
\begin{threeparttable}
\begin{tabular}{c c c c c c c}
\toprule
Dataset & Model & Rouge-1 F1 & Rouge-2 F1 & Rouge-L F1 & BERTscore & Meteor\\

\hline
\multirow{7}*{BookSum Chapter}
& Full article (lower bound) & 13.742 & 4.019 & 13.421 & 0.805 & 21.299\\ 
~ & Lead-3 & 17.683 & 2.747 & 16.708 & 0.812 & 9.815\\ 
~ & Lead-K & 29.149 & 4.641 & 28.034 & 0.805 & 24.091\\
~ & Longformer(baseline) & 33.636 & 9.626 & 32.611 & 0.846 & 27.160 \\
~ & RSTformer(w/o relations) & 33.604 & 10.149 & 32.631 & 0.850 & 26.811\\
~ & RSTformer(w/ relations) & 34.019 & \cellcolor{mypink}\textbf{10.275}$^\dag$$^\ddag$ & \cellcolor{mypink}\textbf{32.870} & \cellcolor{mypink}\textbf{0.853}$^\dag$$^\ddag$ & \cellcolor{mypink}\textbf{27.473}$^\ddag$\\
~ & SOTA model \cite{kryscinski-etal-2022-booksum} & \cellcolor{mypink}\textbf{37.510} & 8.490 & 17.050 & 0.156 & -\\
\hdashline
~ & Our compared to baseline & $+\Delta$0.383 & $+\Delta$0.649 & $+\Delta$0.259 & $+\Delta$0.007 & $+\Delta$0.313\\
~ & Our compared to SOTA & $-\Phi$3.491 & $+\Phi$1.785 & $+\Phi$15.820 & $+\Phi$0.697 & $\Phi$-\\
\midrule
\midrule

\multirow{7}*{eLife}
& Full article (lower bound) & 6.893 & 2.327 & 6.675 & 0.831 & 13.864\\  
~ & Lead-3 & 16.266 & 3.634 & 15.088 & 0.832 & 7.163\\ 
~ & Lead-K & 37.188 & 7.971 & 35.151 & 0.832 & 25.331\\
~ & Longformer(baseline) & 46.778 & 13.318 & 44.317 & \cellcolor{mypink}\textbf{0.855} & 27.921\\
~ & RSTformer(w/o relations) & 46.862 & 14.008 & 44.458 & \cellcolor{mypink}\textbf{0.855} & 27.685\\
~ & RSTformer(w/ relations) & \cellcolor{mypink}\textbf{48.696}$^\dag$$^\ddag$ & \cellcolor{mypink}\textbf{14.843}$^\dag$$^\ddag$ & \cellcolor{mypink}\textbf{46.129}$^\dag$$^\ddag$ & 0.847 & \cellcolor{mypink}\textbf{29.526}$^\dag$$^\ddag$\\
~ & SOTA model \cite{goldsack-etal-2022-making} & 46.570 & 11.650 & 43.700 & - & -\\
\hdashline
~ & Our compared to baseline & $+\Delta$1.918 & $+\Delta$1.525 & $+\Delta$1.812 & $-\Delta$0.008 & $+\Delta$1.605\\
~ & Our compared to SOTA & $+\Phi$2.126 & $+\Phi$3.193 & $+\Phi$2.429 & $\Phi$- & $\Phi$-\\
\midrule
\midrule

\multirow{7}*{Multi-LexSum} 
& Full article (lower bound) & 3.862 & 2.198 & 3.786 & 0.784 & 8.825\\
~ & Lead-3 & 16.135 & 6.387 & 15.421 & 0.770 & 9.538\\
~ & Lead-k & 29.145 & 9.276 & 27.734 & 0.784 & 24.266\\ 
~ & Longformer(baseline) & 45.751 & 21.272 & 43.131 & 0.865 & 33.282\\
~ & RSTformer(w/o relations) & 46.424 & 22.730 &  43.978 & 0.867 & 33.808\\
~ & RSTformer(w/ relations) & 46.421 & 22.888$^\dag$$^\ddag$ & \cellcolor{mypink}\textbf{43.979} & \cellcolor{mypink}\textbf{0.867}$^\ddag$ & \cellcolor{mypink}\textbf{33.941}\\
~ & SOTA model \cite{shen2022multi} & \cellcolor{mypink}\textbf{53.730} & \cellcolor{mypink}\textbf{27.320} & 30.890 & 0.420 & -\\
\hdashline
~ & Our compared to baseline & $+\Delta$0.670 & $+\Delta$1.616 & $+\Delta$0.848 & $+\Delta$0.002 & $+\Delta$0.659\\
~ & Our compared to SOTA & $-\Phi$7.309 & $-\Phi$4.432 & $+\Phi$13.089 & $+\Phi$0.447 & $\Phi$-\\

\bottomrule
\end{tabular}

\end{threeparttable}
}
\caption{Model performance. The bold numbers represent the best results with respect to the given test set. $\Delta$ and $\Phi$ represent the improvement of our model compared to the baseline and SOTA models, respectively. $^\dag$ and $^\ddag$ indicate statistical significance (p$<$0.05) against the baseline model via T-test and Kolmogorov-Smirnov test. Each result of the three distinct SOTA models is directly replicated from their original papers.}
\label{tab:Model Performance}
\end{table*}

\paragraph{Ablation Study}

We also define two additional control conditions to examine the impact of RST attention ($\mathrm{LDD}$) on model performance:

\begin{table}[ht]
\centering
\tabcolsep=3pt
\scalebox{0.75}{
\begin{threeparttable}
\begin{tabular}{c c c c c}
\toprule
Dataset & Model & Rouge-1 & Rouge-2 & Rouge-L\\

\hline
\multirow{2}*{BookSum} 
& Longformer & 33.636 & 9.626 & 32.611\\
~ & RSTformer(WAC) & 31.956 & 8.772 & 31.049\\
Chapter & RSTformer(RIA) & 32.881 & 9.067 & 31.899\\
\hdashline
\multirow{3}*{eLife} 
& Longformer & 46.778 & 13.318 & 44.317\\
~ & RSTformer(WAC) & 39.076 & 8.461 & 37.114\\
~ & RSTformer(RIA) & 41.761 & 10.901 & 40.062\\
\hdashline
\multirow{3}*{Multi-LexSum} 
& Longformer & 45.751 & 21.272 & 43.131\\
~ & RSTformer(WAC) & 42.903 & 18.440 & 40.773\\
~ & RSTformer(RIA) & 42.213 & 20.785 & 31.219\\
\bottomrule
\end{tabular}
\end{threeparttable}
}
\caption{F1 scores for ablation study}
\label{tab:Ablation Study}
\end{table}

\begin{itemize}
    \item \textbf{Without Attention Calculation (WAC)}: We skip the previous calculation of attention weights, and directly replace attention weights with LDD tensor. 
    \item \textbf{Random Identical Attention (RIA)}: We assign fixed random values to LDD tensor, regardless of the probability of discourse relations. 
\end{itemize}

Table \ref{tab:Ablation Study} shows that the RST attention cannot fully replace the calculation of the attention mechanism. Although the performance is significantly lower than the baseline model, its main noteworthy advantage is that it saves considerable computations and parameters. Experiments by introducing random noise demonstrate that random values do indeed negatively impact the model's performance. Furthermore, it also confirms the effectiveness of incorporating the probability distributions of discourse structure.

\paragraph{Human Evaluation}

To better analyze the effectiveness of our model, we randomly select 10 samples from the BookSum dataset and hire human annotators to conduct the human evaluation. The recruited annotators are all master's students or doctoral students with computer science-related or computational linguistics-related backgrounds. All annotators were compensated with the standard hourly salary set by the university. At the time of evaluation, we provide 3 candidate summaries for each source document, namely outputs from our final proposed model and baseline model, along with the ground truth summary. Each instance is assigned to 3 participants who are instructed to rate the faithfulness, informativeness, readability, and conciseness of the candidate summaries on a scale of 1 to 5. They are also supposed to give an overall rank of three summaries and identify which one is generated by humans. Detailed information regarding the human evaluation process can be found in Appendix \ref{sec:appendixB}. Table \ref{tab:Human evaluation} reports the human evaluation results.

\begin{table}[ht]
\centering
\scalebox{0.55}{\tabcolsep=4pt
\begin{threeparttable}
\begin{tabular}{c c c c c c}
\toprule
Candidate & Faithful  & Informative & Readable & Concise &  Best $\mid$ Worst\\

\hline
Human & 4.40 & 4.83 & 4.83 & 4.33 & 83.3\% $\mid$ 0.0\% \\
Longformer & 2.50 & 2.57 & 3.43 & 2.70 & 6.7\% $\mid$ 56.7\% \\
RSTformer(w relations) & 2.97 & 2.90 & 3.73 & 3.00 & 10.0\% $\mid$ 43.7\% \\
\bottomrule
\end{tabular}
\end{threeparttable}
}
\caption{Human evaluation results}
\label{tab:Human evaluation}
\end{table}

For each human evaluation indicator, we compute the average value to represent whether the candidate system has good performance in that indicator. Best and Worst indicate the proportion of times a summary by a particular model is judged to be best or worst among the three options. While neural summarization models still exhibit a notable performance gap when compared to human-generated summaries, our proposed model consistently outperforms the baseline model across all metrics.

\subsection{Analysis}

\begin{figure}[ht]
\centering
\includegraphics[width=0.47\textwidth]{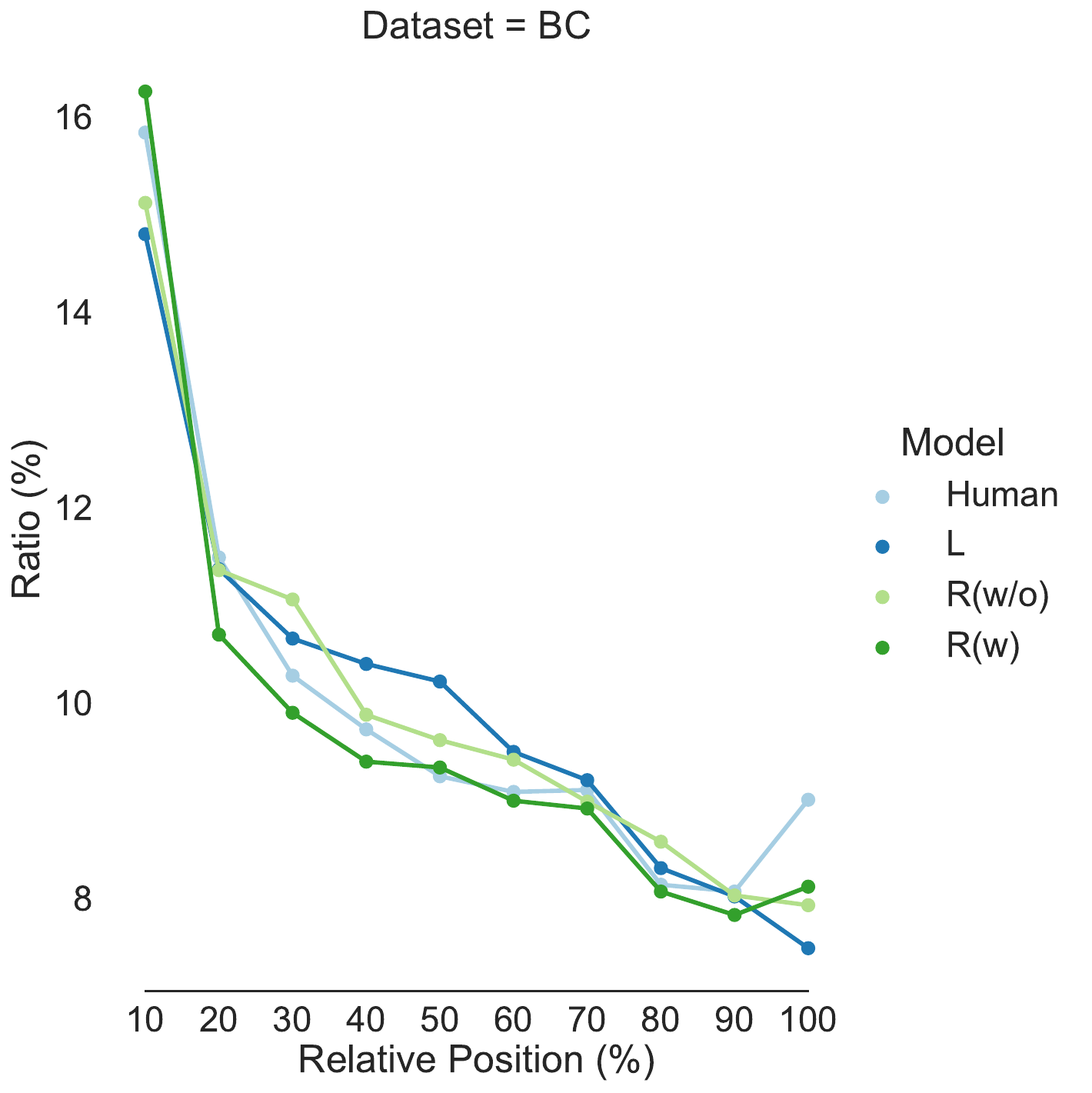}
\caption{Sentence alignment distribution. L = Longformer, R(w/o) = RSTformer(w/o relations), R(w) = RSTformer(w relations), BC = BookSum Chapter.}
\label{fig:Word_Alignment_BC}
\end{figure}

\paragraph{Sentence Alignment}
We examine the alignment distributions of generated summaries to explore whether the improved model can be closer to human-summarized text \cite{liu-liu-2021-simcls}. Our results are depicted in Figure \ref{fig:Word_Alignment_BC} and Appendix \ref{sec:appendixC}.

From a broader perspective, the sentence alignment distribution of our proposed models is more closely aligned with that of human summarizers. In addition, the generated summaries produced by our models demonstrate a greater emphasis on the content of the second half of the document, resulting in summaries that are more comprehensive and coherent in nature.

\paragraph{N-gram Novelty \& Inconsistency Detection}

\begin{figure}[htbp]
\centering
\includegraphics[width=0.49\textwidth]{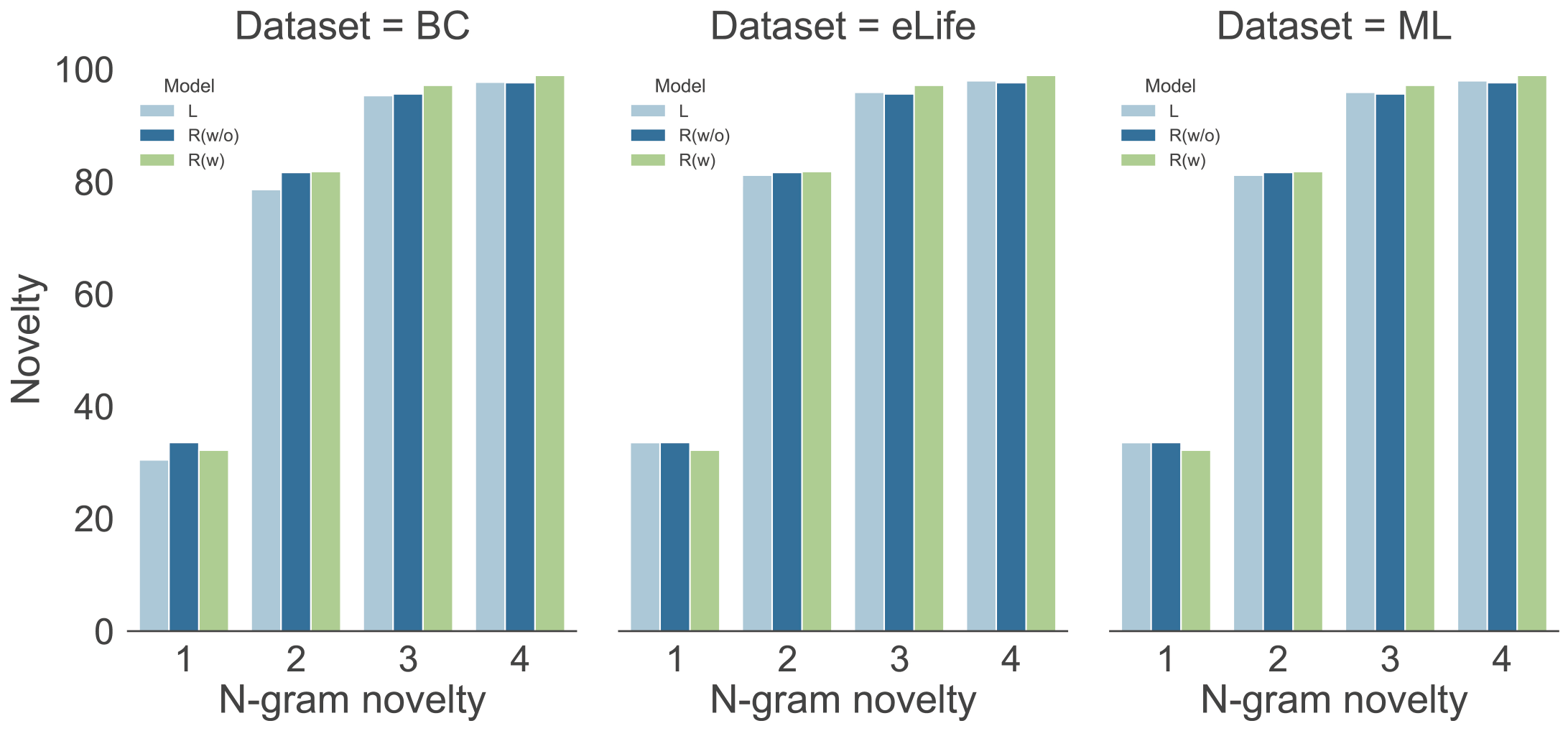}
\caption{N-gram novelty. L = Longformer, R(w/o) = RSTformer(w/o relations), R(w) = RSTformer(w relations), BC = Booksum Chapter, ML = Multi-LexSum.}
\label{fig:n-gram}
\end{figure}

\begin{figure}[ht]
\centering
\includegraphics[width=0.5\textwidth]{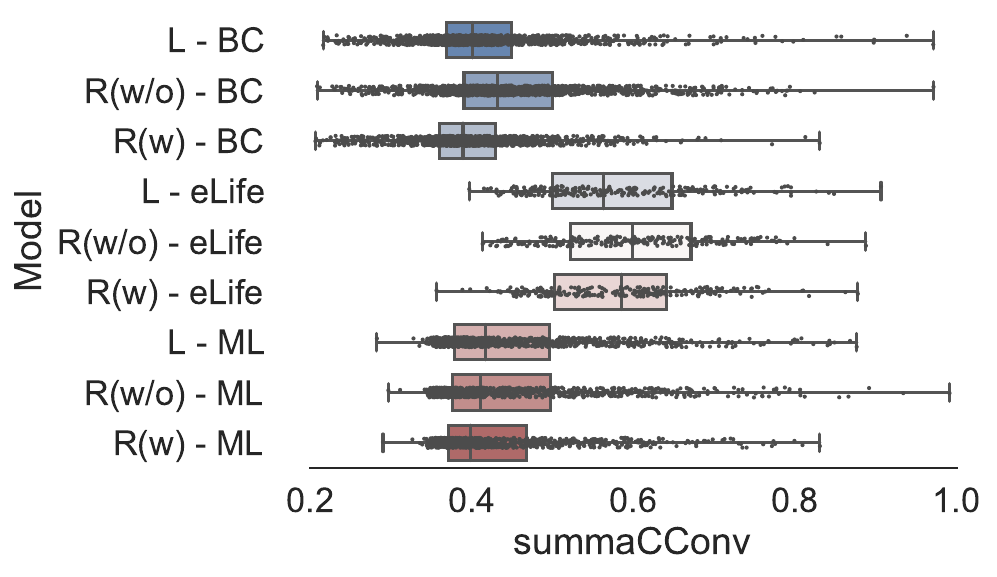}
\caption{Consistency check. L = Longformer, R(w/o) = RSTformer(w/o relations), R(w) = RSTformer(w relations), BC = Booksum Chapter, ML = Multi-LexSum.}
\label{fig:consistency check}
\end{figure}

We also study the level of abstractiveness and factual consistency in the generated summaries. To evaluate the abstractiveness, we employed N-gram novelty as a measure to determine whether the model can generate words that are not present in the original text, rather than solely extracting content from the source document. For inconsistency detection, we utilize the latest SummaC method \cite{laban-etal-2022-summac} for testing. Our results are shown in Figure \ref{fig:n-gram} and Figure \ref{fig:consistency check} respectively.

Compared with the baseline model, incorporating discourse information into the model does increase the ability of the model to generate novel words, especially evident in the context of 3-gram and 4-gram, the gap becomes more prominent. In addition, the proposed model also performs better than the baseline model in terms of model consistency checks. Due to the increased ability to generate creative words, the semantic coherence ability of the models incorporating typed discourse relations is lower than that of models without typed discourse relations.

\section{Conclusion}

This paper introduces a novel supervised discourse enhanced Longformer model. This strategy mainly improves the local attention mechanism in the Longformer model by leveraging the rhetorical structure as uncertainty distributions. The experimental findings provide strong evidence that the proposed approach is straightforward, and can effectively employ the discourse structure of source documents to improve the summary performance of Longformer. Furthermore, this strategy also has a high potential capability for application in other seq2seq natural language tasks.

\section{Limitations}
The present study has certain limitations that should be acknowledged. Firstly, the RST parsing task itself is known to be highly complex and challenging, and achieving high accuracy in this task is not guaranteed. Although we have utilized the most high-performing parser, there is still room for further improvement in the RST parsing performance, which could potentially enhance the downstream summarization task.

Another limitation pertains to the size of the data used for human evaluation. Due to the nature of long document summarization and the length of the original texts (often spanning several pages), scaling up the evaluation process, such as through crowd-sourcing, becomes difficult. Consequently, we are only able to evaluate a limited number of 10 documents, which may not be fully representative of the entire dataset.

Furthermore, another potential risk in our study is the limitation in obtaining an unlimited number of training samples. The data samples investigated are often small subsets of real-world data or may exhibit certain biases, which may not accurately reflect the distribution of real-world data. Although we have verified the effectiveness of our model using highly diverse and heterogeneous datasets from different domains, it is important to note that the model's performance on the specific dataset of interest may not be as robust as its performance on unseen real-world data.

Finally, both training and evaluating the models require significant computational resources. Despite our attempts to optimize the computation by replacing the original attention calculation with the RST attention tensor (as demonstrated in the ablation experiment), we have not achieved satisfactory results. The high computational costs pose a challenge, as they result in increased human and material resources required for the model.

\section{Ethics Considerations}

The datasets we use are all public, and our experiment processes have no privacy disclosure issues. As for human evaluation, all participants are voluntary and paid, and come from master or doctoral students with a background in computer science or computational linguistics, and all of them are proficient in English. They first need to read the instructions and evaluate without revealing which model generates which summary.

\section*{Acknowledgements} 
This project has received funding from the European Research Council (ERC) under the European Union’s Horizon 2020 Research and Innovation Programme (Grant Agreement No. 948878). We are grateful to the anonymous reviewers and area chairs for their exceptionally detailed and helpful feedback.
\begin{figure}[H] 
\centering
\includegraphics[width=0.5\columnwidth]{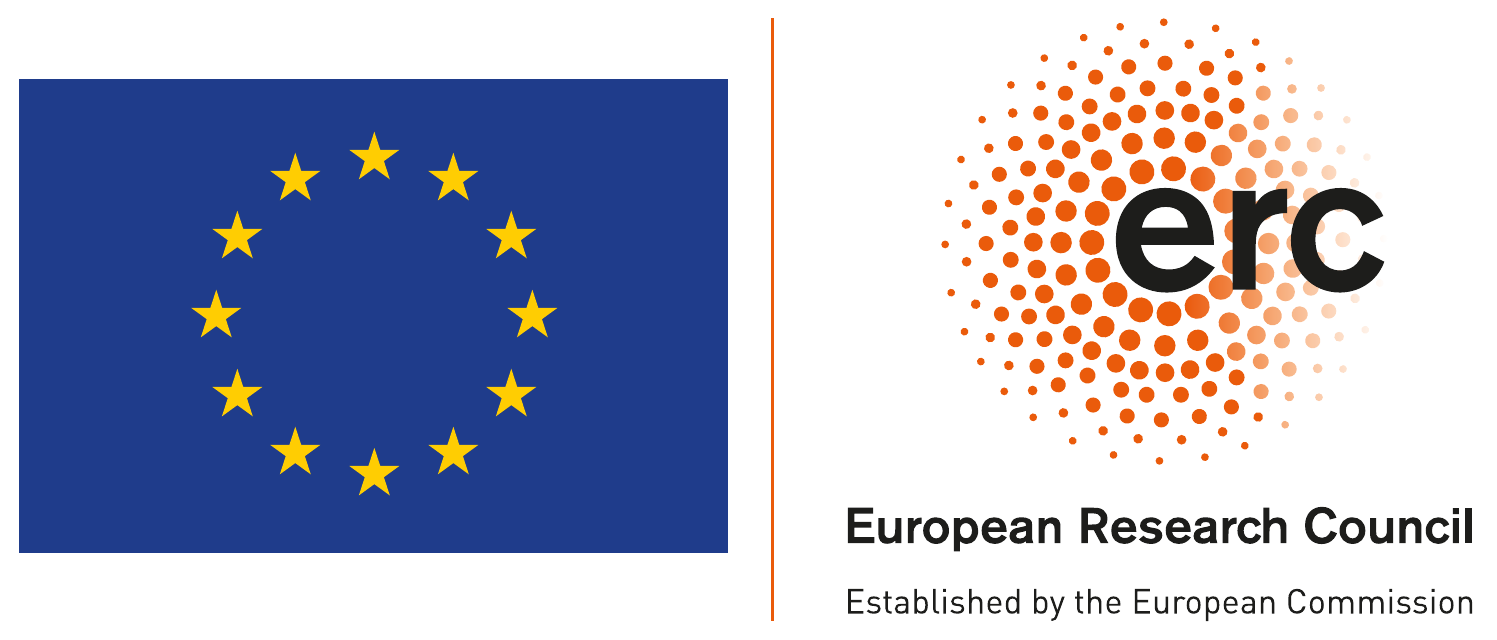}
\end{figure}

\bibliography{anthology,custom}
\bibliographystyle{acl_natbib}

\clearpage

\appendix
\section{Appendix: RST Relation Category}
\label{sec:appendixA}

\begin{table}[!ht]
\centering
\label{tab:RST relation type}
\scalebox{0.7}{
\begin{tabular}{c l}
\toprule
RST type &  RST label \\
\midrule
\textit{Temporal} & Temporal\\
\textit{Contingency} & Cause, Condition\\
\textit{Comparison} & Comparison, Contrast, Concession, Topic-Change\\
\textit{Expansion} & Explanation, Elaboration, Background, Topic-Comment\\
\bottomrule
\end{tabular}
}
\caption*{Table A: RST relation category}
\label{appendix:category}
\end{table}

\section{Appendix: Questionnaire of Human Evaluation}
\label{sec:appendixB}

Here we provide a more detailed description of the criterion in our human evaluation.
\begin{itemize}
    \item \textbf{Faithfulness}
    \begin{enumerate}
        \item Completely hallucinated content
        \item A lot of hallucinated content and factual mistakes
        \item Most content is supported by the source document
        \item Only one or two characters or events contradicted or not mentioned in the source
        \item All information in the summary is faithful/supported by the source
    \end{enumerate}
    \item \textbf{Informativeness}
    \begin{enumerate}
        \item No important information in the source is covered in the summary
        \item Only covers a small fraction of the source document information; one cannot learn the main content of the story from only the summary
        \item Covers around half of the important points from the source; one can learn the main content of the story from only the summary
        \item Only a few important points are missing in the summary
        \item All important information is summarized
    \end{enumerate}
    \item \textbf{Readability}
    \begin{enumerate}
        \item Not understandable at all
        \item Hard to understand the content of the summary
        \item The summary is overall readable, with most sentences correct and fluent
        \item Easy to understand, with only occasional grammatical mistakes or incoherent sentences
        \item Fluent, with minor or no grammatical mistakes, coherent sentences, and clear structure
    \end{enumerate}
    \item \textbf{Conciseness}
    \begin{enumerate}
        \item All information in the summary is redundant or unimportant
        \item Most of the information in the summary is redundant or unimportant
        \item Around half of the content in the summary is redundant
        \item Only a few points in the summary are redundant
        \item No information in the summary is redundant
    \end{enumerate}
\end{itemize}

\begin{figure}[htb]
\centering
\includegraphics[width=0.47\textwidth]{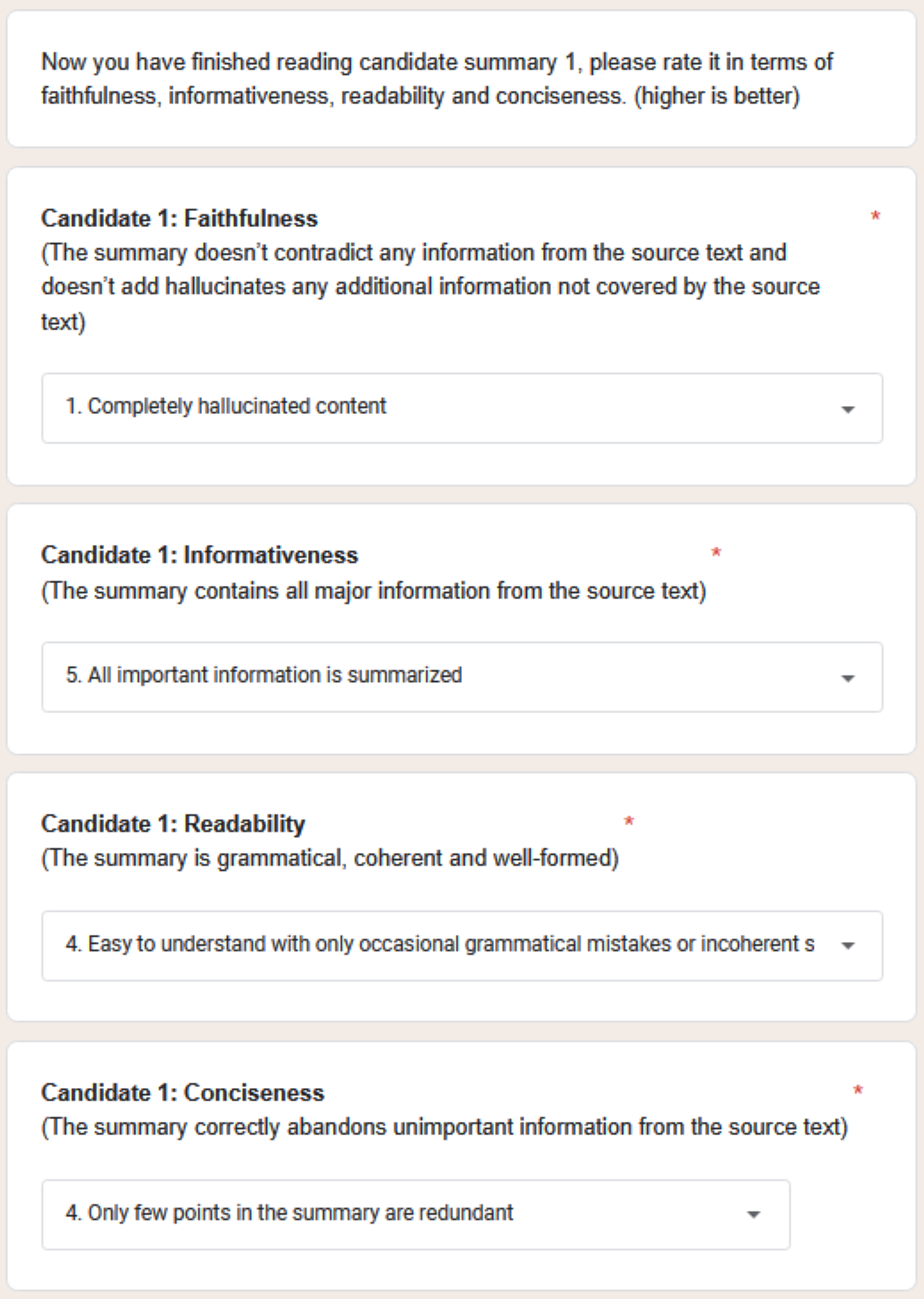}
\caption{Instructions to rate candidate summaries in terms of each metric in human evaluation.}
\label{fig:rating}
\end{figure}

\begin{figure}[htb]
\centering
\includegraphics[width=0.47\textwidth]{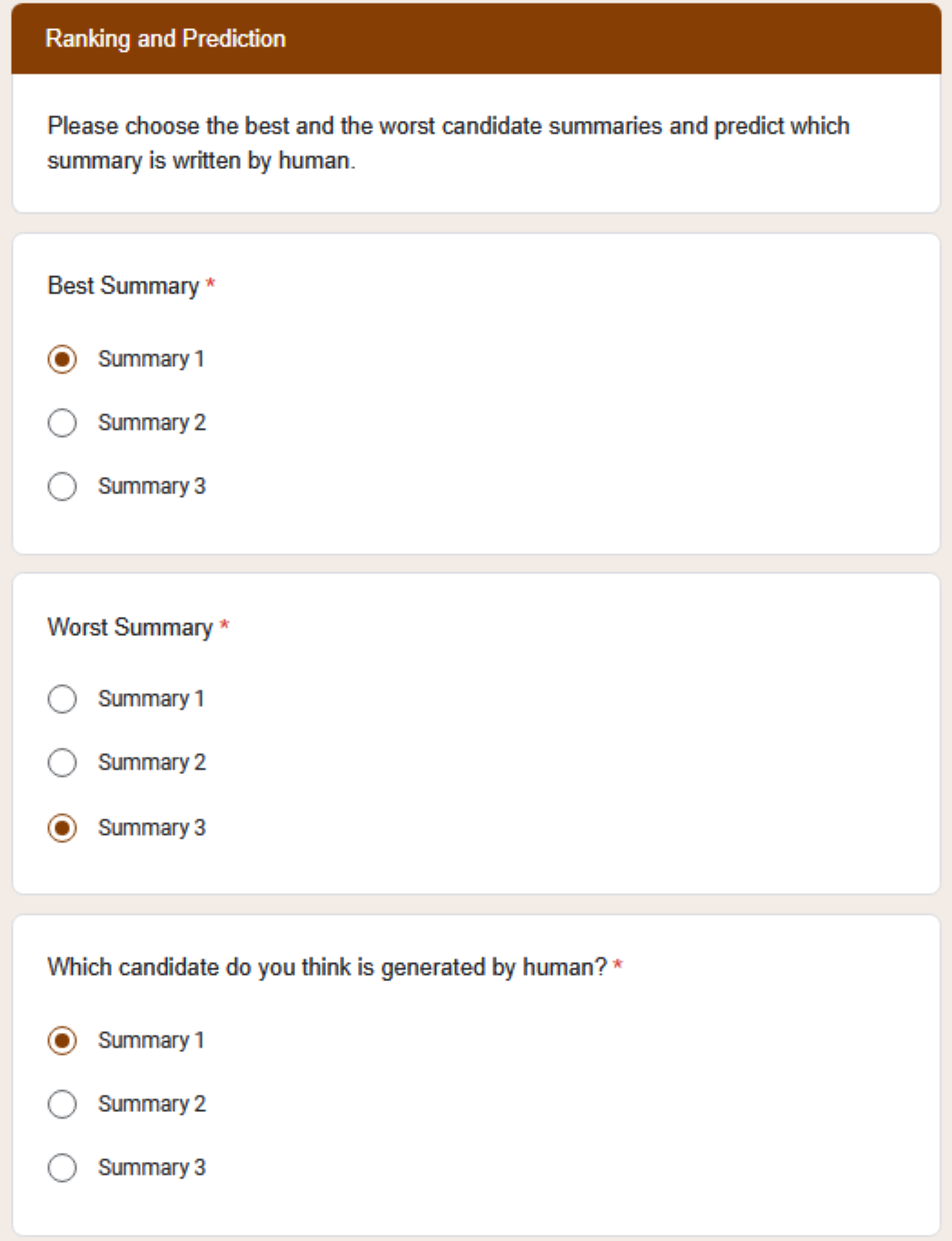}
\caption{Instructions to rank all three candidates and predict which one is generated by human.}
\label{fig:ranking}
\end{figure}

User interface and instructions for rating and ranking can be found in Figure \ref{fig:rating} and Figure \ref{fig:ranking}.

\section{Appendix: Sentence Alignment for Other Datasets}
\label{sec:appendixC}

\begin{figure}[htbp]
\centering
\includegraphics[width=0.47\textwidth]{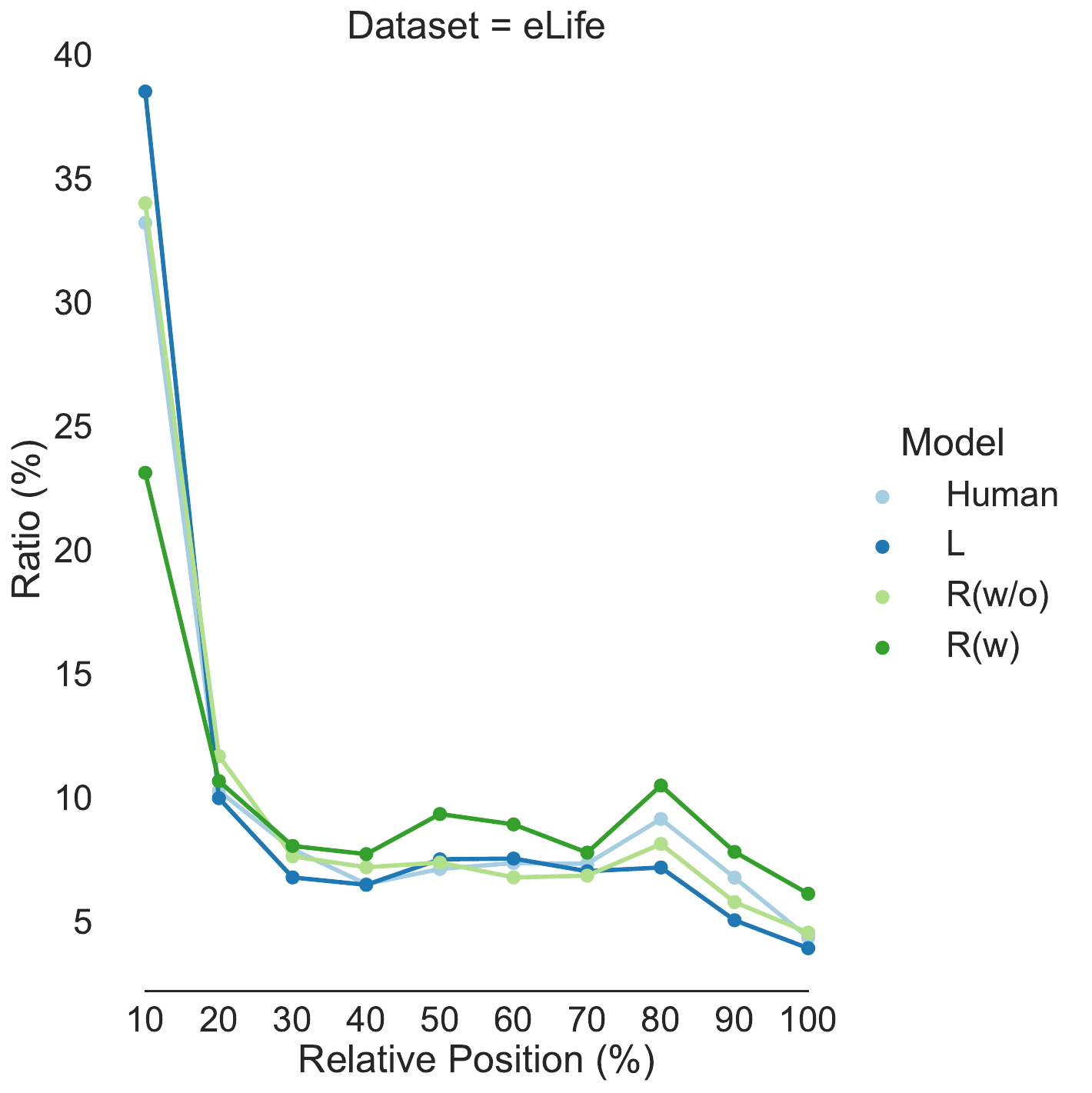}
\caption{Sentence alignment distribution. L = Longformer, R(w/o) = RSTformer(w/o relations), R(w) = RSTformer(w relations).}
\label{fig:Word_Alignment_elife}
\end{figure}

\begin{figure}[htbp]
\centering
\includegraphics[width=0.47\textwidth]{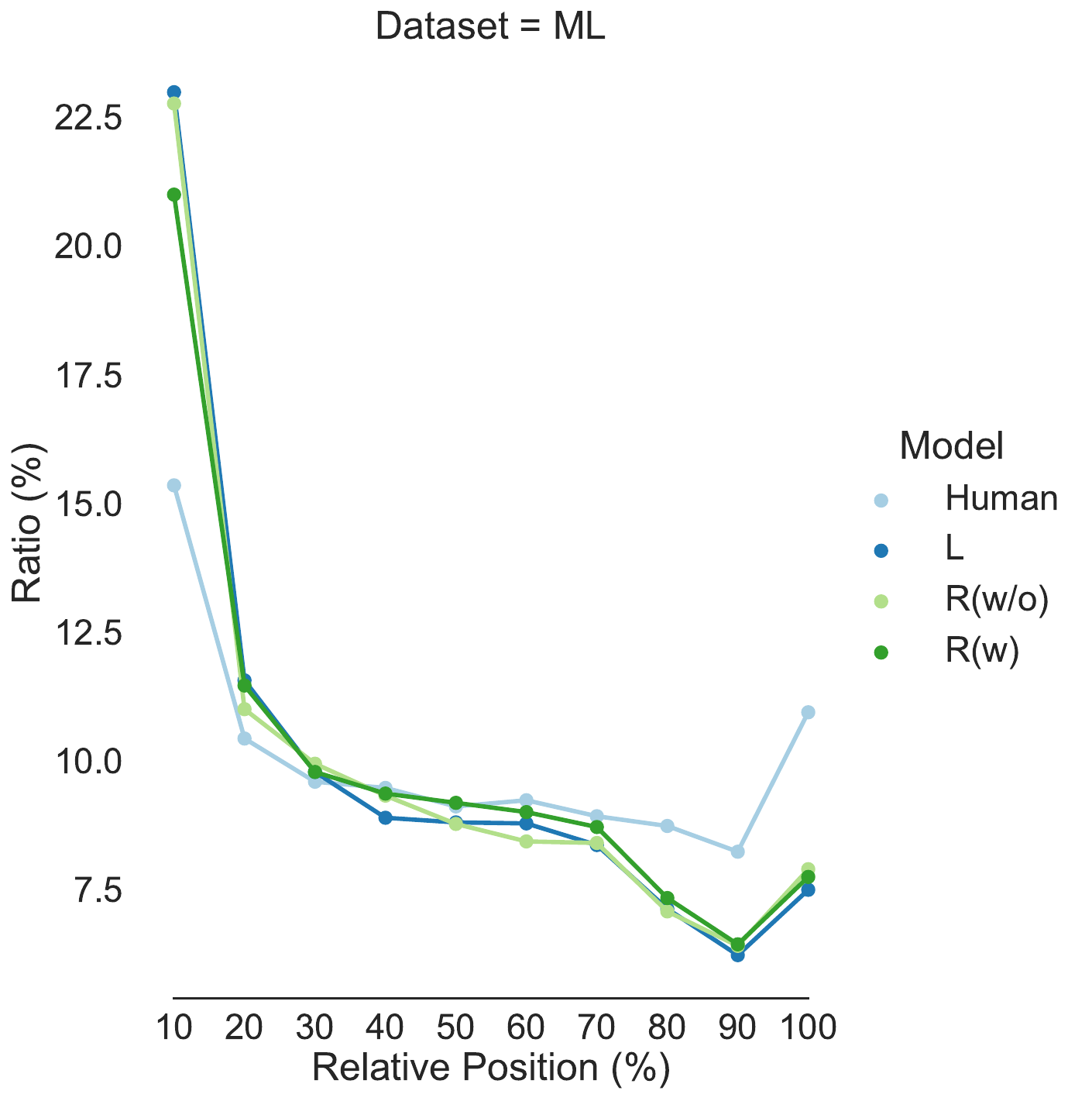}
\caption{Sentence alignment distribution. L = Longformer, R(w/o) = RSTformer(w/o relations), R(w) = RSTformer(w relations), ML = Multi-LexSum.}
\label{fig:Word_Alignment_ML}
\end{figure}

\end{document}